\documentclass{article}

    \PassOptionsToPackage{numbers, compress}{natbib}

\usepackage[final]{neurips_2020}

\usepackage[colorlinks=true]{hyperref}
\usepackage{xspace}
\newcommand{\littleheader}[1]{\noindent \textbf{#1}}
\newcommand{\autoselect}{AutoSelect\xspace}

\usepackage[svgnames]{xcolor}
\usepackage{colortbl}
\usepackage[utf8]{inputenc} 
\usepackage[T1]{fontenc}    
\usepackage{hyperref}       
\usepackage{url}            
\usepackage{booktabs}       
\usepackage{amsfonts}       
\usepackage{nicefrac}       
\usepackage{microtype}      
\usepackage{tikz}
\usepackage{Definitions}
\usepackage{amsmath}
\usepackage[english]{babel}
 \usepackage{array,multirow,graphicx}
 \usepackage{float}
\usepackage{caption}
\usepackage{subcaption}
\usepackage{longtable,tabu}
\usepackage[symbol]{footmisc}
\usepackage[ruled,vlined]{algorithm2e}
\usepackage{wrapfig}
\SetKwComment{Comment}{$\triangleright$\ }{}
\usetikzlibrary{arrows,positioning} 
\tikzset{
    >=stealth',
    punkt/.style={
           rectangle,
           rounded corners,
           draw=black, very thick,
           text width=6.5em,
           minimum height=2em,
           text centered},
    pil/.style={
           ->,
           thick,
           shorten <=2pt,
           shorten >=2pt,}
}
\title{Learning to Select the Best Forecasting Tasks for \\ Clinical Outcome Prediction}

\author{%
   Yuan Xue\textsuperscript{*}, Nan Du\textsuperscript{*}, Anne Mottram, Martin Seneviratne, Andrew M. Dai \\
   Google \\
   \texttt{\{yuanxue, dunan, annemottram, martsen, adai\}}@google.com \\
}

\begin{document}

\maketitle

\begingroup\renewcommand\thefootnote{*}
\footnotetext{Authors contributed equally.}
\endgroup

\begin{abstract}
The paradigm of `pretraining' from a set of relevant auxiliary tasks and then `finetuning' on a target task has been successfully applied in many different domains. However, when the auxiliary tasks are abundant, with complex relationships to the target task, using domain knowledge or searching over all possible pretraining setups is inefficient and suboptimal. To address this challenge, we propose a method to automatically select from a large set of auxiliary tasks, which yields a representation most useful to the target task. In particular, we develop an efficient algorithm that uses automatic auxiliary task selection within a nested-loop meta-learning process. We have applied this algorithm to the task of clinical outcome predictions in electronic medical records, learning from a large number of self-supervised tasks related to forecasting patient trajectories. Experiments on a real clinical dataset demonstrate the superior predictive performance of our method compared to direct supervised learning, naive pretraining and simple multitask learning, in particular in low-data scenarios when the primary task has very few examples. With detailed ablation analysis, we further show that the selection rules are interpretable and able to generalize to unseen target tasks with new data.


\end{abstract}
\section{Introduction}
\vspace{-3mm}


The wide adoption of electronic medical record (EMR) systems has generated large repositories of patient data in the form of multivariate time-series. These data are increasingly used for supervised learning, with the goal of providing decision support to clinicians by predicting clinical outcomes for individual patients~\cite{rajkomar18scalable}. Recent examples have focused on the prediction of inpatient mortality~\cite{Shickel2019}, acute kidney injury~\cite{Tomasev2019}, circulatory shock~\cite{Hyland2020}, etc. 

One major challenge with EMR modeling is that the raw data is high-dimensional, noisy, sparse and heterogeneous, as it is generated in the course of routine clinical care~\cite{Xiao2018}. Furthermore, accurately labeling clinical endpoints can be extremely challenging and often requires time-consuming manual chart review by clinicians, meaning that modeling must be data efficient. Even in cases where the outcome label is more clearly encoded, e.g. mortality, data availability is often still an issue as there are only a limited number of patients with that outcome in a selected cohort.



To tackle these issues of data quality and label shortage, a common approach widely applied in computer vision (CV) and natural language processing (NLP) domains is pretraining and finetuning. Pretraining involves learning a compact representation on related tasks with abundant data. These learned representations can then be finetuned on the primary task with limited labels, assisting supervised performance by leveraging prior knowledge.


EMRs contain thousands of different laboratory tests, observations, medications, procedures \etc, for each patient over time. Using the trajectories of these time series data as self-supervised objectives provides a promising way to learn a useful patient representation. However, naively pretraining across all measurements can easily lead to a representation oblivious of the target clinical prediction task. Furthermore, pretraining from trivial and less important measurements may overshadow important signals in the learned representation. Since the number of available measurement trajectories is large, an exhaustive search over all possible task combinations is not tractable. With complex relationships between the measurement trajectories and the primary outcome, the decision of how to structure pretraining is not a straightforward process. 




To address this challenge, the goal of this paper is to automatically select and mix the most relevant auxiliary tasks to be used in pretraining for a specific primary task. In particular, we introduce a new connection between multitask learning and transfer learning within the framework of meta learning. Each auxiliary task is a self-supervised trajectory forecast for a specific clinical measurement, and the primary target task involves supervised learning based on the learned representation. We propose an efficient gradient-based algorithm that learns to automatically select the most relevant auxiliary tasks for pretraining, and then optimizes the meta objective of generalizing to the target task.

Experiments on real world clinical datasets show that the learned representation from the selected auxiliary tasks leads to favorable predictive performance compared to both direct supervised learning, naive pretraining and simple multitask learning. This advantage further increases in low data regimes where the target task has few labeled examples. Detailed ablation analysis demonstrates that the selected auxiliary tasks are meaningful and able to generalize to unseen target tasks.
\vspace{-3.5mm}

\section{Learning Tasks}
\label{sec:model}
\vspace{-3mm}
In a longitudinal EMR dataset, a patient's record is a collection of sequential clinical-visit data which can be naturally represented as multi-variate time series. Each time series captures the readings over time from one type of clinical measurement (e.g., blood pressure, lactate, etc.), or intervention (e.g., ventilator settings). For a given patient, we use $x^f_t$ to represent the $f$th feature value $f\in\Fcal$ at the time step $t$. $\mathbf{x}^f_T=\{x_t^f\}^T_{t=1}$ denotes the $f$th time series, and $T$ is the number of time steps. We also use $\xb_t$ as the $|\Fcal|$-dimensional feature vector at the time $t$.

\textbf{Primary supervised task: Clinical outcome prediction.} For each sequence $\{\xb_t\}_{t=1}^T$, there is an associated label $y$ representing the occurrence of a clinical outcome of interest, e.g., sepsis, shock, mortality, etc. The goal is to learn a model that predicts the most likely label value $\hat{y}$ for a given input sequence $\{\xb_t\}_{t=1}^T$. The learning process thus takes the standard form of supervised learning with a loss $\ell(\hat{y}, y)$ associated with the model.

\textbf{Auxiliary task: Trajectory forecast.} 
The goal of the trajectory forecast task is to model the distribution of the future values of raw EMR data elements $p(\mathbf{x}^f_{\tau+1:\tau+H}|\mathbf{x}^f_{1:\tau})$ given the past history $\mathbf{x}^f_{1:\tau}$. Here, $\tau$ is the time of prediction, $H$ represents the number of time steps we look into the future, and $\mathbf{x}^f_{\tau+1:\tau+H}=\{x_t^f\}^{\tau+H}_{t=\tau+1}$. This task by nature takes the form of self-supervised learning since the future values of a time series can be easily treated as the learning signal. Compared to the clinical outcome prediction task, the patient's trajectory forecast task requires no human labels, and many powerful self-supervised techniques can be applied to the task~\cite{Dai2015-fv, nips19-attentive-schaar, ml4h19-dssm-xue, nipsw15-dkf, aaai17-sontag-dmm, iclr17-dvbf, nips17-marco-kvae, nips16-marco-srnn}. 

We can expect that pretraining with self-supervised trajectory forecast tasks for each feature $f\in\Fcal$ will produce useful patient representations for the clinical outcome prediction task which often has few examples. However, when the set of auxiliary tasks $|\Fcal|$ is large, both joint pretraining using all the tasks in $\Fcal$, or successive pretraining in an iterative way, can be sub-optimal and inefficient in that not all the auxiliary tasks are equally useful for transferring knowledge to the target primary task, leading to a less informative representation for downstream tasks.

\newcommand{\pth}{p_\theta}
\newcommand{\qph}{q_\phi}
\newcommand{\xfull}{\mathbf{x}_{1:T}}
\newcommand{\zfull}{\mathbf{z}_{1:T}}
\newcommand{\utwofull}{\mathbf{u}_{2:T}}
\newcommand{\ufull}{\mathbf{u}_{1:T}}
\newcommand{\x}{\mathbf{x}}
\newcommand{\z}{\mathbf{z}}
\newcommand{\sumt}{\sum_{t=1}^T}
\newcommand{\sumtm}{\sum_{t=1}^{T-1}}
\newcommand{\xh}{\mathbf{x}_{1:t^*}}
\newcommand{\zh}{\mathbf{z}_{1:t^*}}
\newcommand{\uh}{\mathbf{u}_{1:t^*}}
\newcommand{\xf}{\mathbf{x}_{t^*+1:t^*+\tau}}
\newcommand{\zf}{\mathbf{z}_{t^*+1:t^*+\tau}}
\newcommand{\uf}{\mathbf{u}_{t^*+1:t^*+\tau}}
\newcommand{\sumf}{\sum_{t=t^*+1}^{t^*+\tau}}
\newcommand{\Exp}[2]{\mathop{\mathbb{E}}_{#2}\left[#1\right]}

\begin{figure}
    \centering
    \includegraphics[width=\textwidth]{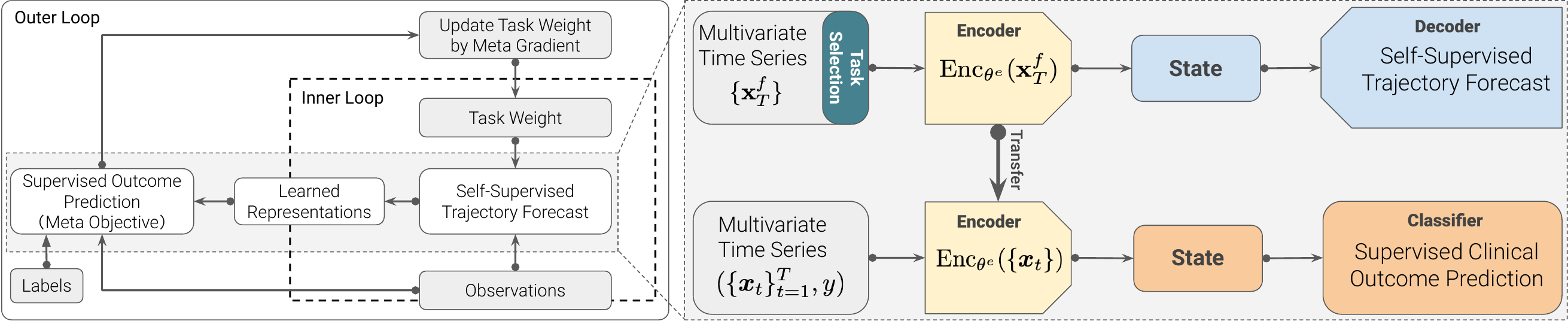}
    \caption{\small{Schematic for guiding pretraining by supervised learning on a primary outcome via a nested-loop meta-learning process. The inner loop learning uses a sequence-to-sequence architecture for trajectory prediction. The utility of the learned representation from the encoder is measured by the supervised learning in the outer loop, and then used to update the weight of each trajectory prediction task in the inner loop via gradient descent. }}
    \vspace{-6mm}
    \label{fig:illustration}
\end{figure}


\section{Automatic Task Selection}
\vspace{-3mm}
We study the problem of learning to select the most relevant trajectory forecast tasks so that the learned representation is optimized for improving the performance of the target clinical outcome prediction task (schematic in Figure~\ref{fig:illustration}). In the following sections, we present our problem formulation, model design, and learning algorithms.

\subsection{Problem Formulation}
\vspace{-2mm}

Selecting the optimal subset of auxiliary trajectory forecast tasks from $\Fcal$ requires exploring $2^{|\Fcal|}$ combinations, which is prohibitive in practice. To make the search space continuous, we relax the categorical choice of a particular task $f$ by learning a weight ${\lambda}_f$, $\sum_f\lambda_f=1$ to indicate the importance of each task. Alternatively, $\lambda$ can be treated as a learned distribution over the task space $\Fcal$. For each task $f\in\Fcal$, without loss of generality, we use a very basic sequence to sequence model, which can easily be extended to more advanced attention-based models~\cite{nips19-attentive-schaar} or state-space models~\cite{ml4h19-dssm-xue, nipsw15-dkf, aaai17-sontag-dmm, iclr17-dvbf, nips17-marco-kvae, nips16-marco-srnn}.
\begin{align}
    \mathbf{Enc}_{\theta^e}(\mathbf{x}^f_{1:\tau})=s_\tau, \,\,  \mathbf{Dec}_{\theta^d}(s_\tau)=\mathbf{\hat{x}}^f_{\tau+1:\tau + H}, \,\,
    \ell^{p}({\theta^e, \theta^d| \lambda})=\sum_f\lambda_f\|\mathbf{\hat{x}}^f_{\tau+1:\tau+H}- \mathbf{x}^f_{\tau+1:\tau+H}\|^2_2,\nonumber
\end{align}
where the encoder $\mathbf{Enc}_{\theta^e}$ encodes the input sequence $\mathbf{x}^f_{1:\tau}$ up to time $\tau$ into an intermediate state $s_\tau$ from which the decoder $\mathbf{Dec}_{\theta^d}$ decodes the predicted sequence $\mathbf{\hat{x}}^f_{\tau+1:\tau+H}$. Here, we use the regression loss for simplicity. $\ell^p$ denotes the pretraining loss. Other reconstruction losses, \eg, negative log likelihood losses, can also be easily applied. For the primary supervised learning task, we reuse the pretrained encoder, and train an additional classifier head for outcome prediction with a typical cross-entropy log loss
\begin{align}
    \mathbf{Enc}_{\theta^e}(\{\xb_{1:\tau}\})=h_\tau,\quad
    \ell^c({\theta^e, \theta^c})=\log{p_{\theta^c}(y|h_\tau)},
\end{align}
where $\theta^c$ are the parameters of the classifier, and $(\{\xb_{1:\tau}\}, y)$ is the training example of one patient, and $\ell^c$ denotes the classification loss.. We formulate the task selection problem into the following optimization task.
\begin{align}
    & \mini_\lambda\ell^c_{\text{val}}\big(\lambda|\theta^{e}_{N_S}, \theta^c_{N_S}\big)\,\, \text{such that}\nonumber\\
    & \theta^{e}_{N_S},\,\theta^c_{N_S}=\argmin\ell^c_{\text{train}}(\theta^e, \theta^c),\,\,\,\theta^{e}_{N_S}=\Phi_{\text{train}}(\theta^{e}_{N_P})
    \xrightarrow{}\text{supervised learning} \nonumber\\
    & \theta^{e}_{N_P}, \theta^{d}_{N_P}=\argmin_{\theta^e, \theta^d}\ell^p_{\text{train}}(\theta^e, \theta^d| \lambda)\xrightarrow{}\text{self-supervised learning}\label{eqn:metaobj}
\end{align}
where $N_P$ and $N_S$ are the number of training update steps of the \textbf{P}retraining and the \textbf{S}upervised training, respectively. $\theta^{e}_{N_P}$ and $\theta^{d}_{N_P}$ are the encoder and decoder parameters learned after $N_P$ steps of self-supervised training. We then switch to the target task by treating $\theta^{e}_{N_P}$ as the initialization of the encoder. $\Phi_{\text{train}}(\theta^{e}_{N_P})$ is an operator that updates the initialization $\theta^{e}_{N_P}$ for $N_S$ steps to obtain $\theta^{e}_{N_S}$. In practice, $\Phi_{\text{train}}$ can be the normal gradient update using the train data of the target task. Here, we explicitly use $\Phi_{\text{train}}$ to denote the initialization from $\theta^{e}_{N_P}$.
More importantly, because the self-supervised learning output $\theta^{e}_{N_P}$ implicitly depends on a given $\lambda$, it further enables the learning of $\lambda$ using the error signals back-propagated from the target task. That is, we explicitly optimize the validation loss $\ell^c_{\text{val}}\big(\lambda|\theta^{e}_{N_S}, \theta^c_{N_S}\big)$ of the target task, which is often referred to as the response function (or meta objective), with respect to $\lambda$ known as the hyper (or meta) parameter, so that the generalization of the target task can be directly optimized.

\subsection{Bi-level Optimization}
\vspace{-2mm}

The optimization of the meta objective~\ref{eqn:metaobj} determines the quality of the learned representation from the encoder with parameter $\theta^e_{N_p}$, and it includes two loops of learning processes shown in Figure~\ref{fig:illustration}. Given a fixed $\lambda$ as one configuration, the inner loop first finds a candidate representation through the self-supervised learning on the trajectory forecast tasks, some of which receives more attention while others may be discarded. It then finetunes the representation via supervised learning on the clinical outcome prediction task. The quality of the learned representation is measured by the meta objective of the outer loop. The outer loop then updates the configuration of the inner loop to locate a potentially better hypothesis space where the two objectives $\ell^p$ and $\ell^c$ will be minimized again. This nested learning process also arises in gradient-based hyper-parameter optimization\cite{icml18-bilevel-luca, icml17-bilevel-luca}, and can be reformulated as follows.
\begin{align}
 &\mini_{\lambda, \cbr{\theta^e_i}, \cbr{\theta^d_i}, \cbr{\theta^c_i}}\ell^c_{\text{val}}\big(\theta^{e}_{N_S}, \theta^c_{N_S}\big)\,\, \text{such that}\nonumber\\
 &\theta^e_i=\Psi^{e}_i(\theta^e_{i-1},\theta^d_{i-1},\lambda),\,\theta^d_i=\Psi^d_i(\theta^e_{i-1},\theta^d_{i-1},\lambda),\,i\in\sbr{1, N_P},\nonumber\\
 &\theta^e_i=\Phi^e_i(\theta^e_{i-1},\theta^c_{i-1}),\,\theta^c_i=\Phi^c_i(\theta^e_{i-1},\theta^c_{i-1}),i\in\sbr{N_P + 1, N_P + N_S},\,\label{eqn:metaobj1}
\end{align}
where $\Psi^\cdot_i$ and $\Phi^\cdot_i$ represent the gradient step of the optimization that updates the parameters at step $i$ in the respective pretrain and finetune stage. This reformulates the implicit dependencies among the parameters in the training procedure into explicit optimization constraints. The Lagrangian of problem~\ref{eqn:metaobj1} is thus
\begin{align}
&\Lcal(\lambda, \cbr{\theta^e_i}, \cbr{\theta^d_i}, \cbr{\theta^c_i},\alpha,\beta,\gamma,\delta)=\ell^c_{\text{val}}\rbr{\theta^{e}_{N_S}, \theta^c_{N_S}}+\sum_{i=1}^{N_P}\alpha_i\rbr{\Psi^e_i(\theta^e_{i-1},\theta^d_{i-1},\lambda) - \theta^e_i} + \\
&\sum_{i=1}^{N_P}\gamma_i\rbr{\Psi^d_i(\theta^e_{i-1},\theta^d_{i-1},\lambda) - \theta^d_i} + \sum_{i=N_P+1}^{N_P +N_S}\beta_i\rbr{\Phi^e_i(\theta^e_{i-1},\theta^c_{i-1}) - \theta^e_i} + \delta_i\rbr{\Phi^c_i(\theta^e_{i-1},\theta^c_{i-1}) - \theta^c_i}\nonumber\end{align}

where for each step $i$, $\alpha_i$, $\beta_i$, $\gamma_i$, and $\delta_i$ are the associated row vectors of Lagrangian multipliers. Since the encoder parameter $\theta^e_{N_p}$ at the last step of the pretraining builds the connection between the self-supervised trajectory forecast tasks and the supervised clinical outcome prediction task, and the value at the last supervised step $\theta^e_{N_s}$ is used for predictions, their derivatives are first given by
\begin{align}
    &\nabla_{\theta^e_{N_P}}\Lcal=-\alpha_{N_P} + \beta_{N_P + 1} \rbr{\nabla_{\theta^e_{N_P}}\Phi^e_{N_P+1}(\theta^e_{N_P},\theta^c_{N_P},\lambda) } \\
    &\nabla_{\theta^e_{N_S}}\Lcal=\nabla_{\theta^e_{N_S}}\ell^c_\text{val}-\beta_{N_S}, \nabla_{\theta^d_{N_P}}\Lcal=-\gamma_{N_P}
\end{align}

Then, at each intermediate step $i$ in the pretrain and finetune stage, the respective derivatives are
\begin{align}
 &\nabla_{\theta^e_i}\Lcal=-\alpha_i + \alpha_{i+1}\nabla_{\theta^e_{i}}\Psi^e_{i+1}(\theta^e_i, \theta^d_i, \lambda),\,\text{for}\,\,i\in\sbr{1, N_P-1},\nonumber\\
 &\nabla_{\theta^e_i}\Lcal=-\beta_i + \beta_{i+1}\nabla_{\theta^e_{i}}\Phi^e_{i+1}(\theta^e_i, \theta^c_i),\,\text{for}\,\,i\in\sbr{N_P+1, N_P+N_S-1},\nonumber\\
 &\nabla_{\theta^d_i}\Lcal=-\gamma_i+\gamma_{i+1}\nabla_{\theta^d_{i}}\Psi^d_{i+1}(\theta^e_i, \theta^d_i, \lambda),\,\text{for}\,\,i\in\sbr{1, N_P-1}\label{eqn:gradient}.
\end{align}

Finally, we can derive the gradient of the hyper-parameter $\lambda$ as
\begin{align}
\nabla_\lambda\Lcal=\sum_{i=1}^{N_P}\rbr{\alpha_i\nabla_\lambda\Psi^e_i(\theta^e_{i-1}, \theta^d_{i-1}, \lambda) + \gamma_i\nabla_\lambda\Psi^d_i(\theta^e_{i-1}, \theta^d_{i-1}, \lambda)}. \label{eqn:metagradient}
\end{align}
The optimal conditions are then obtained by setting each derivative to zero.
\begin{align}
    &\beta_{N_S}=\nabla_{\theta^e_{N_S}}\ell^c_\text{val}\big(\theta^{e}_{N_S}, \theta^c_{N_S}\big) \,\xrightarrow{}\text{supervised objective} \\
    & \beta_i = \beta_{i+1}\nabla_{\theta^e_{i}}\Phi^e_{i+1}(\theta^e_i, \theta^c_i),\,\,i\in\sbr{N_P+1, N_P+N_S-1}\xrightarrow{}\text{supervised learning}\label{eqn:beta}\\
    & \alpha_{N_P} = \beta_{N_P + 1}\nabla_{\theta^e_{N_P}}\Phi^e_{N_P+1}(\theta^e_{N_P},\theta^c_{N_P})\,\xrightarrow{}\text{knowledge transfer}\label{eqn:betaalpha}\\
    & \alpha_i = \alpha_{i+1}\nabla_{\theta^e_{i}}\Psi^e_{i+1}(\theta^e_i, \theta^d_i, \lambda),\,\,i\in\sbr{1, N_P-1}\,\xrightarrow{}\text{self-supervised learning}\label{eqn:alpha}
\end{align}
We first observe that Equation~\ref{eqn:beta} back-propagates the signal from the meta-objective that quantifies the utility of the learned representation from the encoder through the supervised learning process. Equation~\ref{eqn:betaalpha} is the \emph{touching point} of the two learning processes that further transfers this signal back to the self-supervised learning stage. Finally, Equation~\ref{eqn:alpha} distributes the signal to each learning step of the pretraining process. Compared to the encoder that is involved in both pretraining and finetuning, the decoder of the sequence to sequence model is only used in the pretraining stage to serve the self-supervised loss only. As a result, even though the decoder $\theta^d_i$ is involved in $\nabla_\lambda\Psi^e_i$ to measure how fast the gradient of the encoder can change w.r.t $\lambda$, the second order information $\nabla_\lambda\Psi^d_i$ from the decoder itself is not needed to update $\lambda$. This is also verified by the optimality condition that $\gamma_{N_P} = 0$ and $\gamma_i = \gamma_{i+1}\nabla_{\theta^d_{i}}\Psi^d_{i+1}\rbr{\theta^e_i, \theta^d_i, \lambda}$ from Equation~\ref{eqn:gradient}. Therefore, the gradient of $\lambda$ can be solely determined by the signals of both $\alpha$ and $\beta$ from Equation~\ref{eqn:beta} to~\ref{eqn:alpha}. The full algorithm is given in the Appendix.





\subsection{Efficient Gradient-based Learning Algorithm}
\vspace{-2mm}

Exact evaluation of Equation~\ref{eqn:metagradient} is expensive in that $\nabla_\lambda\Psi^e_i(\theta^e_{i-1}, \theta^d_{i-1}, \lambda)$ and $\nabla_{\theta^e_{i}}\Phi^e_{i+1}(\theta^e_i, \theta^c_i)$ include the Jacobian and Hessian matrix of the gradient update operation $\Psi^e_i$ and $\Phi^e_{i+1}$. Motivated by related techniques in~\cite{arxiv18-reptile-openai}, we propose an efficient first-order approximation to Equation~\ref{eqn:metagradient}. More specifically, given that $\theta^{e}_{N_S}=\Phi_{\text{train}}(\theta^{e}_{N_P})=\theta^e_{N_P} + \sum_{i=N_P}^{N_P+N_S - 1}\nabla_{\theta^e_i}\ell^c_{\text{train}}$ in Equation~\ref{eqn:metaobj}, the gradients $\cbr{\nabla_{\theta^e_i}\ell^c_{\text{train}}}$ are treated as constants~\cite{arxiv18-reptile-openai}.
By applying the chain rule with the gradient approximation, we can have
\begin{align}
    \frac{\partial{\ell^c_\text{val}}}{\partial{\lambda}}=\frac{\partial{\ell^c_\text{val}}\rbr{\theta^{e}_{N_S}, \theta^c_{N_S}}}{\partial{\theta^{e}_{N_S}}}\cdot
    \frac{\partial{\theta^{e}_{N_S}}}{\partial{\theta^e_{N_P}}}\cdot
    \frac{\partial{\theta^e_{N_P}}}{\partial{\ell^p_{\text{train}}(\theta^e,\theta^d|\lambda)}}\cdot
    \frac{\partial{\ell^p_{\text{train}}(\theta^e,\theta^d|\lambda)}}{\partial{\lambda}},\label{eqn:firstorder}
\end{align}
where we have $\partial{\theta^e_{N_S}}/\partial{\theta^e_{N_P}}$ to be the identity matrix due to the gradient approximation, $\partial{\theta^e_{N_P}}/\partial{\ell^p_{\text{train}}(\theta^e,\theta^d|\lambda)}=1 / \frac{\partial{\ell^p_{\text{train}}(\theta^e,\theta^d|\lambda)}}{\partial{\theta^e_{N_P}}}$ which can be simply achieved at the end of the self-supervised training, and $\partial{\ell^p_{\text{train}}(\theta^e,\theta^d|\lambda)}/\partial{\lambda}$ can be obtained via back-propagation. The overall first-order approximation algorithm is given in Algorithm~\ref{alg:algorithm}. After the joint training, there will be a final round of finetuning on the target task alone. Experimentally, we find this stage is useful when the target task has very few examples, and its contribution decreases as more training examples become available.

\begin{algorithm}[t]
\small
\caption{First-Order Automatic Task Selection}\label{alg:algorithm}
Randomly initialize $\theta^e_0$, $\theta^d_0$, $\theta^c_{N_P}$ and $\lambda$\;
\For{$k = 1, 2, ...$}
{
    \For(\Comment*[f]{Self-supervised learning loop}){$ i \in [1, N_P]$}
    {
        $\theta^e_i=\Psi^e_i\rbr{\theta^e_{i-1}, \theta^d_{i-1}, \lambda_k}$, $\theta^d_i=\Psi^d_i\rbr{\theta^e_{i-1}, \theta^d_{i-1}, \lambda_k}$
        \Comment*[r]{Gradient descent}
    }
    Get $a = {\partial{\ell^p_{\text{train}}}}/{\partial{\theta^e_{N_P}}}$, and $b = {\partial{\ell^p_{\text{train}}}}/{\partial{\lambda}}$\;
    \For(\Comment*[f]{Supervised learning loop}){$ i \in [N_P +1, N_P+N_S]$}
    {
        $\theta^e_i=\Phi^e_i\rbr{\theta^e_{i-1}, \theta^c_{i-1}}$, $\theta^c_i=\Phi^c_i\rbr{\theta^e_{i-1}, \theta^c_{i-1}}$\Comment*[r]{Gradient descent}
    }
    Get $c = {\partial{\ell^c_{\text{val}}}}/{\partial{\theta^e_{N_S}}}$\;
    Get $g_\lambda= c \cdot(1/a) \cdot b$ \Comment*[r]{Compute hyper-gradient by Equation~\ref{eqn:firstorder}}
    $\lambda_k=\lambda_{k-1} - \epsilon\cdot g_\lambda$ \Comment*[r]{Gradient descent}
}
\Return{$\theta^e_{N_P}$, $\lambda$}

\end{algorithm}

\vspace{-3mm}
\section{Experiments}
\vspace{-2mm}
We evaluate our proposed algorithm, referred to as \textbf{\autoselect}, using the openly accessible MIMIC-III dataset~\cite{johnson16mimic} which contains over 38,000 adult patients admitted to the intensive care unit.
We select a set of $96$ common clinical measurements, which constitutes the set of candidate auxiliary tasks used for trajectory forecast. All values were normalized using z-score, and missing values were imputed by carrying forward the last observation. Yet, the model is always trained only using the true values as the targets instead of the imputed values.
\begin{wraptable}{r}{0pt}
\scriptsize
\begin{tabular}{cc}
\toprule 
\rowcolor{GhostWhite}
{\bf Task} & {\bf Definition}  \\
\midrule                
Mortality & Patient expired \\ 
Low Blood Pressure (BP) &	Mean blood pressure $\leq 65$mmhg \\
Kidney Dysfunction (KD) & Creatinine $\geq2$mg/dl	 \\
\bottomrule
\end{tabular}
\caption{\small{Task definitions.}\label{tab:task}}
\vspace{-3mm}
\end{wraptable}
We consider three primary supervised learning tasks defined using the criteria in Table~\ref{tab:task}. The prediction uses data from the first $48$ hours of the ICU admission, and the label is positive if the criteria are fulfilled within the next $48$ hour window (i.e. $48-96$ hours post admission). Moreover, the event sequence of each patient is also restricted to a window of 48 hours in the past, so that the bias towards longer stays can be alleviated.
For simplicity, the latter two organ failure tasks are defined in a lightweight manner following the SOFA score criteria~\cite{sofa_score}. We report the details of the inclusion and exclusion criteria, the cohort and feature statistics, data preprocessing methods and results on additional tasks in the Appendix. 

\subsection{Baselines and Experiment Setting}
\vspace{-2mm}


\littleheader{Supervised Learning.} We train a single baseline model with exactly the same architecture as the model used for the primary tasks of Table~\ref{tab:task} in \autoselect. Given that these primary tasks often have low resources, we expect supervised learning to have low predictive performance in general.

\littleheader{Pretraining (All).} This is the same pretraining-and-finetuning paradigm as in the work of ~\cite{ml4h18-ts-brown}. We first learn the patient representation by pretraining using all the 96 self-supervised trajectory forecast tasks, and then finetune the model on the target tasks in Table~\ref{tab:task}.

\littleheader{CoTrain via Multitask Learning.} This is a simple widely used multitask learning setup. We first cotrain the target task with all the auxiliary tasks, and use a task weight hyperparameter to balance the losses between these two groups of tasks. We set the weight of the target loss to be 10 tuned by the validation set performance to make the losses in the same scale, and the auxiliary task has a weight of 1. After the co-training stage, we finetune the model using only the data of the primary task.

\littleheader{Experimental Setup.} The sequence-to-sequence architecture of the trajectory forecast task uses an LSTM for both encoder and decoder, where the hidden state has a dimension of 70. For the primary clinical outcome prediction task, the decoder is replaced by a simple 1-layer MLP as the classification head. All the baselines and our method use the same architecture. This does not prevent our method from using more advanced architectures. All models were implemented\footnote{\url{https://github.com/google-health/records-research/meta-learn-forecast-task}} in TensorFlow \cite{tensorflow2015-whitepaper}.

For the direct supervised learning baseline, we use early stopping with the validation set to avoid overfitting. For all the experiments of the other approaches, we run approximately 5,000 steps during the pretraining stage, followed by 5 epochs for finetuning. For \autoselect, these 5,000 steps are further divided between inner loop steps and outer meta-learning steps, so that the total number of training steps is consistent with other pretrained methods for fair comparison. The learning rates of all training loops were tuned and are 0.001 for supervised learning, 0.005 for self-supervised learning, 0.01 for $\lambda$ hyper-gradient update. Detailed hyperparameter tuning process is reported in the Appendix. Finally, we use 10-fold cross validation and estimate the standard error of the mean. For each fold, we split the dataset into train/validation/test according to $80\%$/$10\%$/$10\%$ based on the hash value of the patient ID, and AUC-ROC is used as the evaluation metric by default with standard error reported in the parentheses next to it.

\subsection{Performance Comparison of Clinical Outcome Prediction}

\begin{table}[t]
    \centering
    \footnotesize
    \scriptsize
    \renewcommand\arraystretch{1.2}
    \renewcommand\tabcolsep{10pt}
    \begin{tabular}{crcccc}
    \toprule 
    \rowcolor{GhostWhite}
    {\bf{Task}} & {\bf{Data}} & {\bf{Supervised}} & {\bf{Pretrain (All)}} & {\bf{CoTrain}} & {\bf{AutoSelect}}\\
    \midrule 
   \parbox[h]{2mm}{\multirow{3}{*}{\rotatebox[origin=c]{90}{Mortality}}} 
    & 1\% & 0.738 (0.017) & 0.809 (0.010) & 0.725 (0.014) & \cellcolor{Ivory}{\bf 0.833} (0.017)\\
    & 10\% & 0.853 (0.016) & 0.853 (0.013) & 0.854 (0.014) & \cellcolor{Ivory}{\bf 0.882} (0.012)\\
    & 100\% & 0.899 (0.008) & 0.899 (0.011) & 0.902 (0.009) & \cellcolor{Ivory}{\bf 0.909} (0.008)\\
    \midrule 
\parbox[h]{2mm}{\multirow{3}{*}{\rotatebox[origin=c]{90}{BP}}}
    & 1\% & 0.730 (0.022) & 0.778 (0.031) & 0.718 (0.041) & \cellcolor{Ivory}{\bf0.838} (0.022)\\
    & 10\% & 0.754 (0.040) & 0.772 (0.028) & 0.724 (0.031) & \cellcolor{Ivory}{\bf0.833} (0.018)\\
    & 100\% & 0.886 (0.026) & 0.881 (0.030) & 0.892 (0.018) & \cellcolor{Ivory}{\bf0.899} (0.021)\\
    \midrule
\parbox[h]{2mm}{\multirow{3}{*}{\rotatebox[origin=c]{90}{KD}}}
    & 1\% & 0.745 (0.015) & 0.771 (0.021) & 0.748 (0.020) & \cellcolor{Ivory}{\bf0.823} (0.018)\\
    & 10\% & 0.849 (0.015) & 0.828 (0.012) & 0.849 (0.012) & \cellcolor{Ivory}{\bf0.862} (0.018)\\
    & 100\% & 0.901 (0.011) & 0.907 (0.007) & 0.899 (0.009) & \cellcolor{Ivory}{\bf0.910} (0.011)\\
    \bottomrule
    \end{tabular}
    \vspace{2mm}
    \caption{\small{Predictive performance (AUC-ROC) of different competing methods for the three primary outcome prediction tasks under consideration with respect to different levels of data-scarcity.}
    \label{tab:performance}}
    \end{table}
 Table~\ref{tab:performance} shows the results of gradually adding more training data by taking 1\%, 10\% and 100\% from the original train dataset. We first observe that the performance of all methods in all tasks increases as more training data from the primary task are used. In the low resource regime, `Pretrain (All)' has better performance than naive supervised learning, which is expected since it can transfer knowledge by learning from the auxiliary trajectory forecast tasks. Second, we also observe that the `CoTrain' baseline has a hard time to balance all the 96 self-supervised trajectory forecasts and the supervised outcome prediction task even if it has an additional finetuning process. More sophisticated mixing ratios are thus needed to reconcile the different training speed of each task. Finally, we further compare \autoselect to the two-stage pipeline approach~\cite{naacl19-autosem-unc} in the extreme case of using 1\% of the data where it achieves 0.751(0.013), 0.720(0.022), 0.760(0.025) on the task of Mortality, BP and KD, respectively. By comparison, \autoselect learns to adaptively tune the weight of each auxiliary task guided by the validation error from the primary task during pretraining, and thus is able to outperform these baselines by a significantly large margin.

\subsection{Ablation Study}
\vspace{-2mm}

\begin{table}[t]
    \centering
    \renewcommand\tabcolsep{3pt}
    \parbox[t]{.6\linewidth}{
    \vspace{0pt}
    \scriptsize
    \caption{\small{Predictive performance of \autoselect in selected tasks.}\label{tab:ablation}}
    \begin{tabular}{lccccc}
    \toprule 
    \rowcolor{GhostWhite}
    {\bf{Task}} & {\bf{Data}} & {\bf{AutoSelect}} & {\bf{Pretrain (Top)}} & {\bf{Pretrain (Down)}} & {\bf{Pretrain (All)}} \\
    \midrule 
    \multirow{2}{*}{BP} & 1\% & 0.838 (0.022) & \cellcolor{LemonChiffon}0.812 (0.014) & \cellcolor{LightCyan}0.788 (0.019) & 0.778 (0.031) \\
    & 10\% & 0.833 (0.018) & \cellcolor{LemonChiffon}0.824 (0.021) & \cellcolor{LightCyan}0.781 (0.027) & 0.772 (0.028) \\
    \midrule
    \multirow{2}{*}{KD} & 1\% & 0.823 (0.018) & \cellcolor{LemonChiffon}0.805 (0.016) & \cellcolor{LightCyan}0.749 (0.028) & 0.771 (0.021) \\
    & 10\% & 0.862 (0.018) & \cellcolor{LemonChiffon}0.855 (0.021) & \cellcolor{LightCyan}0.825 (0.021) & 0.828 (0.012) \\
    \midrule
    \multirow{2}{*}{Mortality} & 1\% & 0.833 (0.017) &\cellcolor{LemonChiffon} 0.810 (0.013) & \cellcolor{LightCyan}0.772 (0.019) & 0.809 (0.010) \\
     & 10\% & 0.882 (0.012) & \cellcolor{LemonChiffon}0.850 (0.011) & \cellcolor{LightCyan}0.823 (0.014) & 0.853 (0.013) \\ 
    \bottomrule
    \end{tabular}
    }\hfill
    \parbox[t]{.35\linewidth}{
    \renewcommand\tabcolsep{5pt}
    \vspace{0pt}
    \scriptsize
    \caption{\small{Generalization of \autoselect}\label{tab:generalization}}
    \begin{tabular}{ccc}
    \toprule 
    \rowcolor{GhostWhite}
    {\bf{Data}} & {\bf{Mortality$\rightarrow$ BP}} & {\bf{Mortality$\rightarrow$ KD}}\\
    \midrule
    1\% &  0.842 (0.019) & 0.833 (0.017) \\
    10\% &  0.847 (0.020) & 0.869 (0.019) \\
    \midrule
    \rowcolor{GhostWhite}
    \textbf{Data}& {\bf{BP$\rightarrow$ Mortality}} & {\bf{KD$\rightarrow$ Mortality}}\\
    \midrule
    1\% &  0.812 (0.020) & 0.809 (0.018)\\
    10\% & 0.871 (0.012) & 0.867 (0.013) \\
    \bottomrule
    \end{tabular}    
        }
    \vspace{-4mm}
\end{table}

\begin{figure}
    \centering
    \renewcommand\tabcolsep{2pt}
\begin{tabular}{ccc}
     \begin{subfigure}[t]{0.4\textwidth}
        \includegraphics[width=\textwidth]{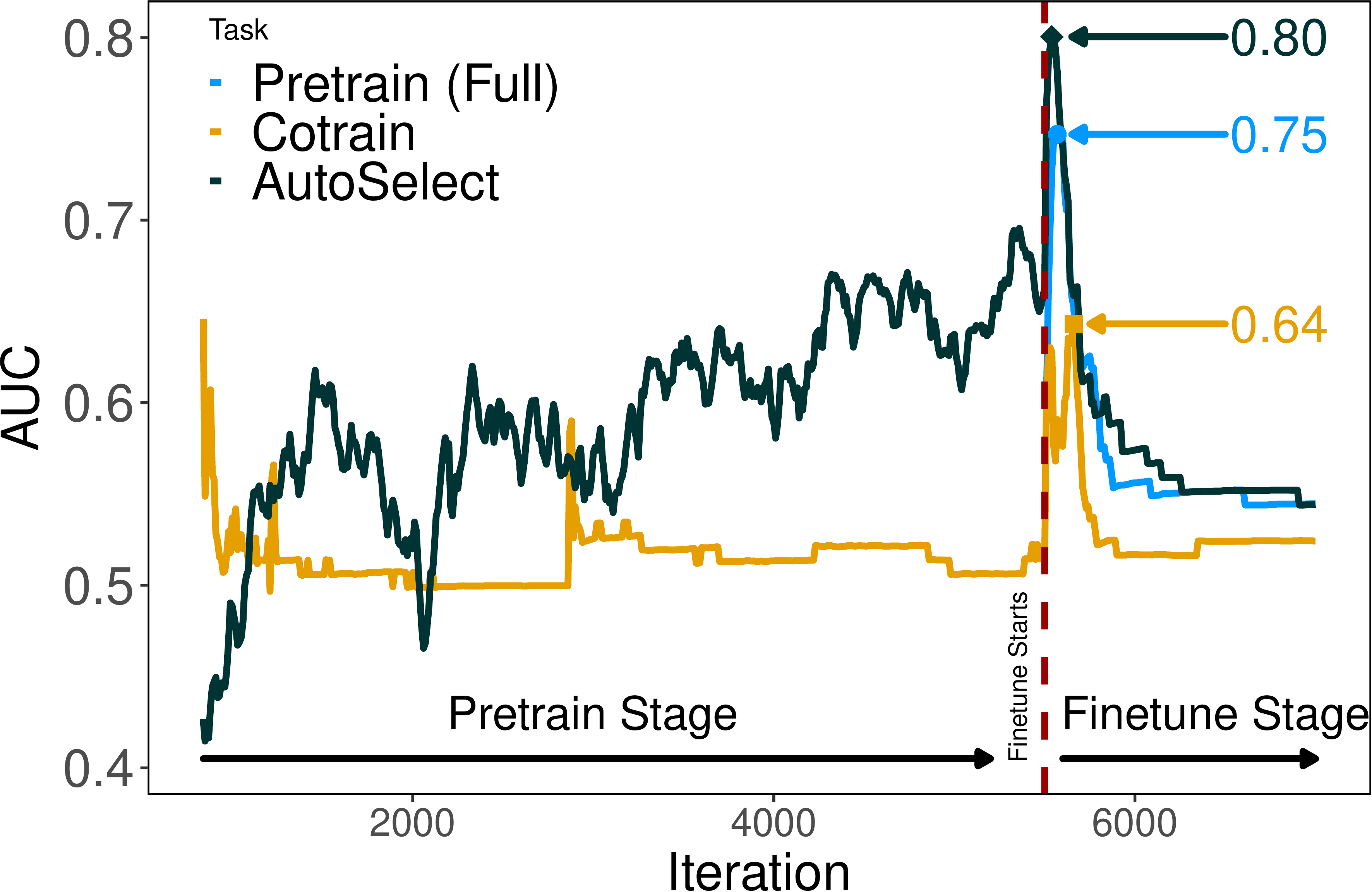}
        \caption{\small{Learning Dynamics}\label{td}}
     \end{subfigure}
     &
     \begin{subfigure}[t]{0.26\textwidth}
        \includegraphics[width=\textwidth]{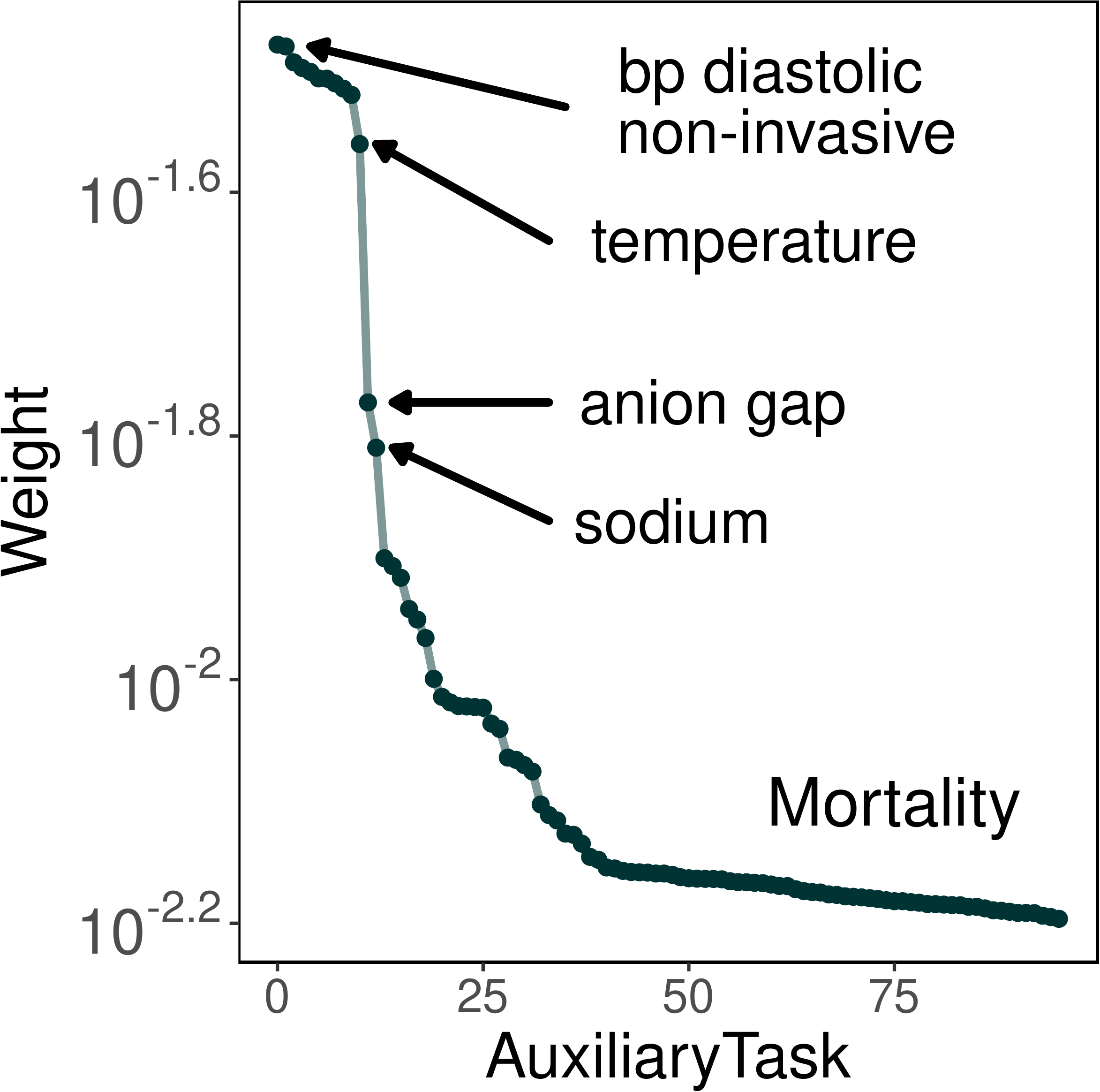}
        \caption{\small{Task Weight Distribution}\label{attention}}
     \end{subfigure}      
     &
     \begin{subfigure}[t]{0.26\textwidth}
        \includegraphics[width=\textwidth]{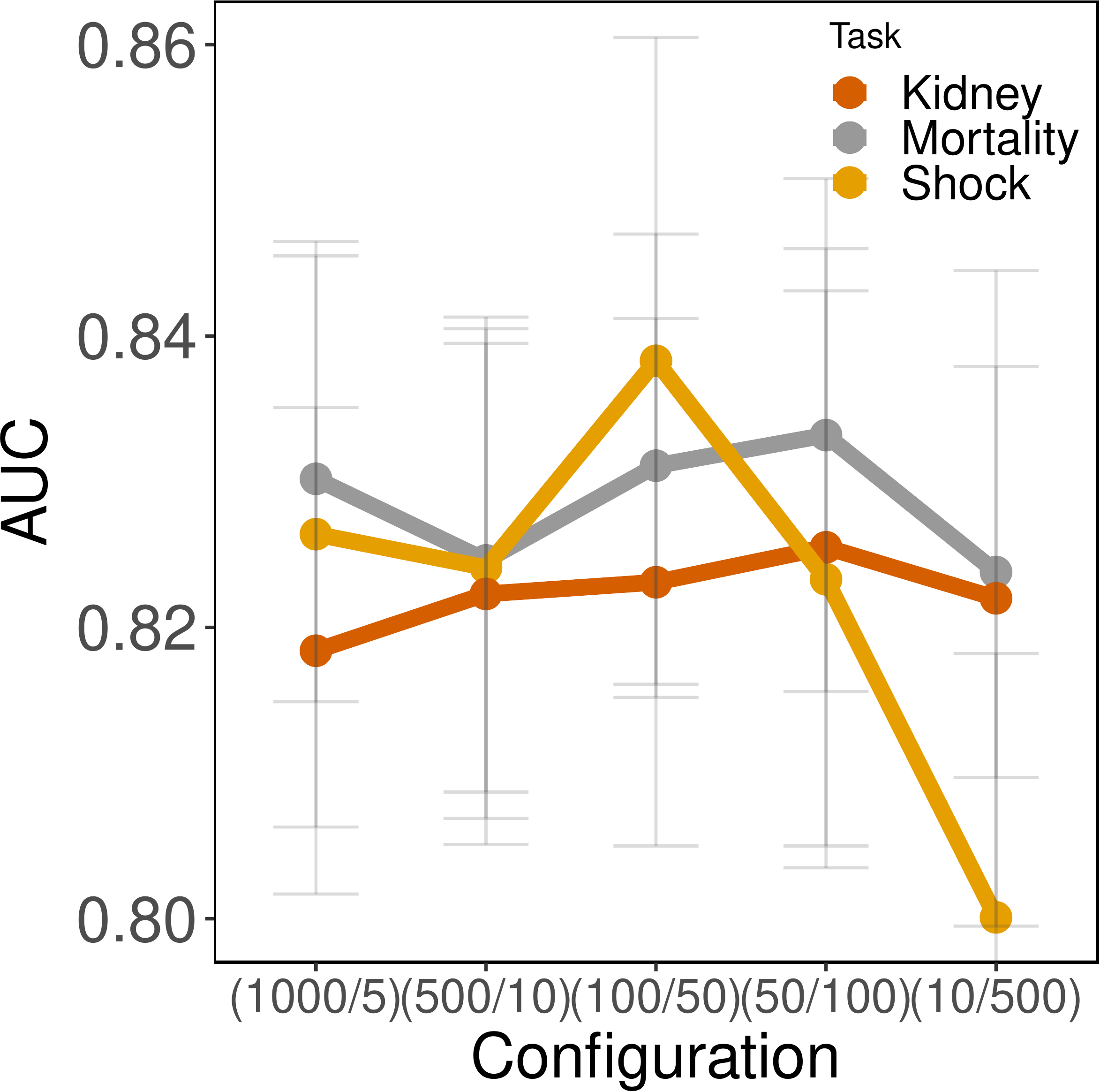}
        \caption{\small{Robustness}\label{hyper}}
     \end{subfigure}
\end{tabular}
\caption{\small{(a) Pretraining and finetuning curves of competing methods. AUC-ROC on the validation set is reported. (b) The learned weights of the 96 auxiliary tasks for mortality prediction as the primary task. 
(c) Predictive performance with respect to different meta-learning processes of \autoselect.} \label{fig:ablation}}
\vspace{-6mm}     
\end{figure}

\littleheader{What tasks are selected?} We now examine the pretraining tasks that were assigned with higher weights in the meta learning process shown in Figure~\ref{attention}. The following features were consistently ranked within the top 20 across different training data splits for mortality prediction: invasive and non-invasive blood pressures, heart rate, anion gap, respiratory rate (Full list is available in the Appendix). These represent a mixture of common vital signs and laboratory values that are clinically sensible correlates for mortality. Indeed, there is significant overlap with the input features for classical risk scores that have been validated as mortality predictors in intensive care (e.g. APACHE II~\cite{Capuzzo2000}). 
The top features for the other two supervised tasks, kidney dysfunction and low blood pressure, are detailed in the Appendix. Notably, the top features for low blood pressure include all available blood pressure recordings; however in the kidney dysfunction task, creatinine, which is the laboratory value on which the outcome is defined, does not appear in this top list. Our hypothesis is that creatinine is measured sparsely - typically once every 24 hours - thus providing a weak signal over the 48 hour window of the self-supervising trajectory forecast task.


\littleheader{How good are the selected tasks?} To further validate the quality of top selected tasks and evaluate the impact of these features as pretraining tasks on the supervised outcome, we have conducted two ablation studies. In the first, we pretrain the encoder using the top selected auxiliary tasks only, referred to as `Pretrain (Top)'. In the second, we instead pretrain the model with the remaining auxiliary tasks excluding the top ones, referred to as `Pretrain (Down)'. The 
hypothesis is that the top selected tasks already capture the necessary knowledge needed to optimize the performance of the target task, while the remaining ones are less important. We report the results of these two studies in Table~\ref{tab:ablation}. It shows that `Pretrain (Top)' performs closer to \autoselect and is consistently better than `Pretrain (Down)' and `Pretrain (Full)', suggesting that the top selected tasks are able to transfer the most useful information to the target task. 
Meanwhile,
we also observe that `Pretrain (Down)' has similar performance to `Pretrain (Full)' showing that the useful signals are indeed overshadowed in the learned representation with the full set of tasks.

\littleheader{How does the learning occur?} Figure~\ref{td} presents the training dynamics of \autoselect for the mortality task as an example. During the pretraining stage, its performance measured by AUC-ROC in the validation set keeps improving and then jumps even higher when the finetuning stage starts at step 5,500 when the validation performance of the auxiliary tasks reaches the peak. The performance of mortality prediction then quickly decreases due to overfitting to its small finetuning data. The blue curve represents the learning process of `Pretrain (All)'. Because the mortality task was not involved in the pretraining stage, it only starts from the beginning of finetuning. The yellow curve is the process of `CoTrain', where the validation performance of the mortality task first jumps and then decreases during the pretraining period. This is caused by the learning speed difference among all the tasks where the training of the small primary task starts to overfit while that of the other auxiliary tasks still improves. Finally, we study the impact of different training steps of the nested learning loops of \autoselect. By fixing the total number of iterations around 5,000, in Figure~\ref{hyper}, we explore different configurations by varying the number of self-supervised training iterations ($N_P$) at the inner loop from 1,000 steps to 10 steps where (1,000/5) means 1,000 inner loop iterations and 5 outer loop iterations. In addition, the inner supervised training loop ($N_S$) is configured to take $1/10$ of the self-supervised iterations ($N_P$). We observe \autoselect is generally robust across different configurations, and sweet points seem to be around (100/50) and (50/100).

\littleheader{How does \autoselect generalize?} Finally, we look at how well the learned weights of the auxiliary tasks guided by a given target task are able to generalize to unseen new tasks. We first use the mortality prediction task as the given primary task to guide the selection of the auxiliary trajectory forecast tasks. Then, we treat BP and KD as two new tasks, and directly finetune the model using 1\% and 10\% of the train data respectively. The reason is that, from a clinical perspective, mortality is a more general endpoint than specific dysfunctions. The hypothesis is that representations learned on the more general endpoint of mortality will be useful for more specific tasks. The prediction results are given in the top row of Table~\ref{tab:ablation}. Despite training only on the mortality task, the learned weights of the auxiliary tasks are able to improve the performance of BP and KD compared to the `Pretrain (All)' baseline. Next, we use BP and KD to guide the selection learning separately, and then test on the mortality task at the bottom of Table~\ref{tab:ablation}. Since the specific dysfunction does not always imply an endpoint of mortality, the predictive performance on mortality is slightly worse than the respective results of \autoselect.


\vspace{-2mm}
\section{Related Work}
\vspace{-3mm}

\textbf{Multi-Task Learning and Task selection}. Recent research on multitask learning~\cite{arxiv17-multitask-survey} aims at either generalizing from a source multitask learning domain to a target multitask domain~\cite{NIPS2018_7819}, or learning a good trade-off among different tasks~\cite{NIPS2019_9374}. There have been approaches to the task scheduling problem via a separate two-stage pipeline~\cite{naacl19-autosem-unc, zaremoodi2020learning}, and~\citet{nips19-cnap-cambridge} learns to adjust model architectures to automatically adapt to new tasks. \citet{iccv2017-multitask-zisserman} combines multiple self-supervised tasks to train useful visual representations. Being complementary, our method learns to automatically select a set of auxiliary tasks, each of which is a self-supervised time-series learning problem, so that pretraining on these tasks can lead to a better representation for a target supervised learning task in an end-to-end differentiable framework.

\textbf{Meta learning}.
Meta learning\cite{arxiv20-meta-survey, iclr17-maml-hugo, icml17-maml-levine, kdd19-metapred-fei} seeks to acquire an initialization optimized for a set of tasks from the same distribution~\cite{nips19-curricula-levine, iclr19-meta-levine, iclr19-meta-brain, kdd19-metapred-fei}. One challenge of meta learning frameworks is that they rely on manually-defined training tasks, and hand-crafting these tasks can be time-consuming. The work of \cite{nips19-curricula-levine} presents unsupervised methods for inducing an adaptive meta-training task distribution and~ the work of \cite{iclr19-meta-levine} automatically constructs tasks from unlabeled data. We address this challenge via  automatic selection over a large number of self-supervised tasks. Our method is similar in spirit to the work of \cite{iclr19-meta-brain} in that both direct the meta-learning process using a supervised target task, but we differ in that \cite{iclr19-meta-brain} meta-learns an unsupervised learning rule, while our work meta-learns a self-supervised task weight distribution.

\textbf{Patient Representation Learning}. The self-supervised task in our work is related to recent progress on state representation learning especially via patient trajectories~\cite{nips19-attentive-schaar, ml4h19-dssm-xue, nipsw15-dkf, aaai17-sontag-dmm}. The objectives in these works are often reconstruction errors in the entire observation space, and thus are not incentivized to capture latent factors that are useful for downstream tasks. To address this issue, the work of \cite{nips18-cpc-deepmind, nips19-stdim-bengio} learn state representations by predicting the future in latent space with a probabilistic contrastive loss, while our work directs the representation learning by reducing error on a downstream target task.

\vspace{-2mm}
\section{Conclusion}
\vspace{-3mm}

We demonstrate how to leverage trajectory forecasts over clinical observations as self-supervised pretraining tasks to improve the quality of clinical outcome predictions. We 
present an efficient algorithm for automatic task selection and primary task training within a nested-loop meta-learning process. Experiments on a real clinical dataset show that our architecture achieves superior predictive performance, in particular in low-data scenarios when the primary task has very few examples.
 

\newpage
\section{Broader Impact}
\vspace{-2mm}

This work presents a method for efficiently learning patient representations using EMR data. Although this is demonstrated with a subset of the full raw EMR, and for only a handful of clinical outcomes in intensive care patients, it is a proof-of-concept that may be useful for a range of other predictive modeling using various types of longitudinal health data. The impact may be greatest in low-data scenarios - e.g. clinical use-cases where labeling is very challenging or where there are few eligible patients in the EMR. The code for this method will be made available to the research community on GitHub. 

There are numerous ethical considerations associated with any EMR modeling, which have been discussed in the literature~\cite{Verheij2018,rajkomar_hardt}. Issues include numerous biases in the observational EMR data, e.g. on the basis of gender, ethnicity or socioeconomic status, which can propagate into predictive models. These fairness considerations also apply to representation learning architectures as presented here.

Finally, if this method were to be brought forward to real world deployment in conjunction with a decision support tool, it would have to be subject to appropriate clinical safety review and trials across different populations, with consideration given to issues such as drift and robustness. 



\bibliography{main}
\bibliographystyle{unsrtnat}
\newpage
\section{Appendix}
\vspace{-2mm}

\subsection{Full Algorithm}

Based on Equation~\ref{eqn:beta} to~\ref{eqn:alpha}, the gradient of $\lambda$ (Equation~\ref{eqn:metagradient}) can be solely determined by the signals of both $\alpha$ and $\beta$. The full algorithm is given in Algorithm~\ref{alg:algorithm-full}.

\begin{algorithm}
\small
\caption{Full Algorithm for Automatic Task Selection}\label{alg:algorithm-full}
Randomly initialize $\theta^e_0$, $\theta^d_0$, $\theta^c_{N_P}$ and $\lambda$\;
\For{$k = 1, 2, ...$}
{
    \For(\Comment*[f]{Self-supervised learning loop}){$ i \in [1, N_P]$}
    {
        $\theta^e_i=\Psi^e_i\rbr{\theta^e_{i-1}, \theta^d_{i-1}, \lambda_k}$, $\theta^d_i=\Psi^d_i\rbr{\theta^e_{i-1}, \theta^d_{i-1}, \lambda_k}$
        \Comment*[r]{Gradient descent}
        $\mathbf{H}^e_i \gets \nabla_{\theta^e_{i}}\Psi^e_{i+1}(\theta^e_i, \theta^d_i, \lambda)$
        \Comment*[r]{Record Hessian Matrix}
        $\mathbf{J}^e_i \gets \nabla_\lambda\Psi^e_i(\theta^e_{i-1}, \theta^d_{i-1}, \lambda)$ \Comment*[r]{Record Jacobian for $\lambda$ over Gradient}
    }
    \For(\Comment*[f]{Supervised learning loop}){$ i \in [N_P +1, N_P+N_S]$}
    {
        $\theta^e_i=\Phi^e_i\rbr{\theta^e_{i-1}, \theta^c_{i-1}}$, $\theta^c_i=\Phi^c_i\rbr{\theta^e_{i-1}, \theta^c_{i-1}}$\Comment*[r]{Gradient descent}
         $\mathbf{H}^e_i \gets \nabla_{\theta^e_{i}}\Phi^e_{i+1}(\theta^e_i, \theta^c_i)$ \Comment*[r]{Record Hessian Matrix}
    }
    $\beta_{N_S} \gets \nabla_{\theta^e_{N_S}}\ell^c_\text{val} $  \Comment*[r]{Initialize multiplier}
\For(\Comment*[f]{Reverse steps of supervised loop}){$  i \in [ N_P+N_S-1, N_P +1]$}{
  $\beta_{i} \gets \beta_{i+1} \mathbf{H}^e_i$ \Comment*[r]{Update multiplier}
}
 $ \alpha_{N_P} = \beta_{N_P + 1} \mathbf{H}^e_{N_P}$ \Comment*[r]{Transfer Supervised to Self-supervised}
 $g \gets 0$ \Comment*[r]{Initialize hyper-gradient}
\For(\Comment*[f]{Reverse steps of self-supervised loop}){$ i \in [N_P, 1]$}  {
 $g_\lambda \gets g_\lambda + \alpha_{i+1} \mathbf{J}^e_{i+1}$ \Comment*[r]{Update hyper-gradient}
$\alpha_{i} \gets \alpha_{i+1}  \mathbf{H}^e_i$ \Comment*[r]{Update multiplier}
}
$\lambda_i=\lambda_{i-1} - \epsilon\cdot g_\lambda$ \Comment*[r]{Hyper-gradient descent}
}
\Return{$\theta^e_{N_P}$, $\lambda$}
\end{algorithm}

Note that the gradient descent steps follow the following equations:
\begin{align}
&\Psi^e_i(\theta^e_{i-1}, \theta^d_{i-1}, \lambda) =  \theta^e_{i-1} - \eta^p \nabla_{\theta^e} \ell^p_{\text{train}}(\theta^e, \theta^d, \lambda) \\
&\Phi^e_i(\theta^e_{i-1}, \theta^c_{i-1}) =  \theta^e_{i-1} - \eta^c \nabla_{\theta^e} \ell^c_{\text{train}}(\theta^e, \theta^c)
\end{align}
where $\eta^p$ and $\eta^c$ are the learning rates for the self-supervised training and supervised training respectively. As thus, $\nabla_\lambda\Psi^e_i(\theta^e_{i-1}, \theta^d_{i-1}, \lambda)$, $\nabla_{\theta^e_{i}}\Psi^e_{i+1}(\theta^e_i, \theta^d_i, \lambda)$ and $\nabla_{\theta^e_{i}}\Phi^e_{i+1}(\theta^e_i, \theta^c_i)$ include the Jacobian and Hessian matrix of the gradient update operation $\Psi^e_i$ and $\Phi^e_{i+1}$ and are cached during the learning loops. Also to simplify notations, we overloaded $N_S$ and use $\beta_{N_S}$ and $\nabla_{\theta^e_{N_S}}$ to represent $\beta_{N_S+N_P}$ and $\nabla_{\theta^e_{N_S+N_P}}$ in the algorithm. 

\subsection{Dataset and Data Preprocessing}

We evaluate our proposed algorithm using the open access MIMIC-III dataset~\cite{johnson16mimic} which contains patients admitted to the intensive care unit of Beth Israel Deaconess Medical Center between 2001 and 2012.

\begin{table}[h!]
\centering
\small
\renewcommand\tabcolsep{2pt}
\caption{Task definition and statistics.\label{table:label_stat}}
\begin{tabular}{ccccc} 
\toprule 
\rowcolor{GhostWhite}
{\bf Task} & {\bf Definition}  &  {\bf Positive}  &  {\bf Negative} & {\bf Excluded}  \\
\midrule                
Mortality & patient expired & 597&	36,773	&1,115\\ 
Shock &	mean blood pressure $\leq 65$&		3,418	&15,159&	19,908 \\
Kidney Failure &	creatinine $\geq2$mg/dl	&	526 &	32,055	& 5,904 \\
Liver Failure &	bilirubin $\geq2$mg/dl	&	387 &	34,856&	3,242 \\
\bottomrule
\end{tabular}
\end{table}

There are 38,485 adult inpatient encounters included in the study. Supervised tasks were triggered at 48 hrs after ICU admission with a lookahead horizon of 48 hours (i.e. 48-96 hours post admission) for the following endpoints: inpatient mortality, renal and liver failure, and circulatory shock. For this proof of concept work, the latter three organ failure endpoints were defined in a lightweight manner based on SOFA score criteria~\cite{sofa_score}, as described in Table~\ref{table:label_stat}, rather than using more detailed definitions taking into account baseline values.

The positive (negative) columns are the number of encounters where the corresponding events are observed (unobserved) after the prediction time (i.e., 48 hrs after ICU admission) and within the prediction window (i.e. 48-96 hours post admission). The excluded column counts the encounters where the corresponding event occurs before the prediction time. We only include the encounters greater than 48 hours in length and where the outcome of interest did not occur within the first 48 hrs.

We select the 96 most frequently occurring lab measurements, vital signs and interventions/equipment settings as features to define the trajectory forecast tasks, as listed in Table~\ref{tab:features}. As different coding systems are used in MIMIC-III, we harmonize the medical codes corresponding to the same lab/vital measurement as a single feature. In addition, we standardize the units when a medical code is used with multiple units or without a unit. The data preprocessing steps are summarized as follows:
\begin{enumerate}
    \item {\bf Code harmonization}. This is a manual process based on input from clinical experts. In this step, the medical codes corresponding to the same measurements from different coding systems, including LOINC and MIMIC specific coding, are harmonized into the same entity. For example, serum creatinine is associated with the following MIMIC-III specific codes: 220615, 50912, 1525, 3750, 791, and LOINC code 2160-0. 
    \item {\bf Unit conversion}. This is an automated process with manual review. In MIMIC-III, a medical code may be stored in multiple units and sometimes the unit might be missing. To determine whether these two entries correspond to the same measurement concept, we derive its value range and mean under different units and test whether they are similar to each other. The final results are reviewed manually. 
    \item {\bf Outlier removal}. We derive the distribution of each measurement code after harmonization and unit conversion. We remove the outliers, defined as $<0.1 \times$ the value at the 1st percentile or $>10 \times$ the value at the 99th percentile. 
    \item {\bf Value normalization}. We collect the mean and standard deviation over the cleaned dataset for each harmonized code and compute its z-score as feature value. 
    \item {\bf Time bucketing}. Each timestep in the experiment corresponds to 1 hour. If multiple measurements exist within this 1 hour for the same clinical feature, we take its average value. 
\end{enumerate}

In Table~\ref{tab:features}, the first column is the name and unit of the feature after harmonization. The second column includes all the names being used for this feature in the MIMIC-III dataset. The corresponding MIMIC-specific and LOINC codes are listed in the third and fourth columns. The `Units' column lists all the original units associated with this feature. We use `null' to denote the case where unit is left empty in the dataset. Finally, the last column lists the number of appearances of this feature in the dataset. 

\begin{table}[h!]

\centering
\small
\renewcommand\tabcolsep{2pt}
\caption{Hyperparameter Selection.\label{tab:hyper-tune}}
\begin{tabular}{p{3cm}cp{3.5cm}c} 
\toprule 
\rowcolor{GhostWhite}
{\bf Hyperparameter} & {\bf Values/Range Considered}  &   {\bf Selection Criteria}  & {\bf Value Selected}  \\
\midrule                
Learning rate for supervised learning & \{0.0005, 0.001, 0.05, 0.01\} & Validation AUC for supervised training and fine-tune after pretraining	& 0.001\\ 
\midrule                
Learning rate for self-supervised learning & \{0.0005, 0.001, 0.05, 0.01\} & Validation MSE in pretraining & 0.005 \\
\midrule                
Learning rate for $\lambda$ update & \{0.001, 0.01, 0.1\}	& Validation AUC for fine-tune after pretraining & 0.01 \\
\midrule                
State size for LSTM	& Range [20, 100] with increment of 10 & Validation MSE in pretraining and validation AUC for supervised training  &	70 \\
\bottomrule
\end{tabular}
\end{table}

\subsection{Hyperparameters and Selections}

The learning rates of all training loops were tuned jointly with the state size of LSTM via a grid search. Table~\ref{tab:hyper-tune} shows the list of values considered, the criteria we use in choosing the hyperparameter and the final value selection.


\begin{table}[t]
    \centering
    \footnotesize
    \renewcommand\arraystretch{1.2}
    \begin{tabular}{crcccc}
    \toprule 
    \rowcolor{GhostWhite}
    {\bf{Task}} & {\bf{Data}} & {\bf{Supervised}} & {\bf{Pretrain (All)}} & {\bf{CoTrain}} & {\bf{AutoSelect}}\\
    \midrule 
   \parbox[h]{2mm}{\multirow{3}{*}{\rotatebox[origin=c]{90}{Mortality}}} 
    & 1\% & 0.738 (0.017) & 0.809 (0.010) & 0.725 (0.014) & \cellcolor{Ivory}{\bf 0.833} (0.017)\\
    & 10\% & 0.853 (0.016) & 0.853 (0.013) & 0.854 (0.014) & \cellcolor{Ivory}{\bf 0.882} (0.012)\\
    & 100\% & 0.899 (0.008) & 0.899 (0.011) & 0.902 (0.009) & \cellcolor{Ivory}{\bf 0.909} (0.008)\\
    \midrule 
   \parbox[h]{2mm}{\multirow{3}{*}{\rotatebox[origin=c]{90}{Shock}}} 
    & 1\% & 0.730 (0.022) & 0.778 (0.031) & 0.718 (0.041) & \cellcolor{Ivory}{\bf0.838} (0.022)\\
    & 10\% & 0.754 (0.040) & 0.772 (0.028) & 0.724 (0.031) & \cellcolor{Ivory}{\bf0.833} (0.018)\\
    & 100\% & 0.886 (0.026) & 0.881 (0.030) & 0.892 (0.018) & \cellcolor{Ivory}{\bf0.899} (0.021)\\
    \midrule
   \parbox[h]{2mm}{\multirow{3}{*}{\rotatebox[origin=c]{90}{Kidney F.}}}
    & 1\% & 0.745 (0.015) & 0.771 (0.021) & 0.748 (0.020) & \cellcolor{Ivory}{\bf0.823} (0.018)\\
    & 10\% & 0.849 (0.015) & 0.828 (0.012) & 0.849 (0.012) & \cellcolor{Ivory}{\bf0.862} (0.018)\\
    & 100\% & 0.901 (0.011) & 0.907 (0.007) & 0.899 (0.009) & \cellcolor{Ivory}{\bf0.910} (0.011)\\
    \midrule
   \parbox[h]{2mm}{\multirow{3}{*}{\rotatebox[origin=c]{90}{Liver F.}}}
    & 1\% & 0.721 (0.017) & 0.780 (0.010) & 0.707 (0.027) & \cellcolor{Ivory}{\bf0.822} (0.013)\\
    & 10\% & 0.863 (0.015) & 0.832 (0.009) & 0.832 (0.009) & \cellcolor{Ivory}{\bf0.866} (0.016)\\
    & 100\% & 0.896 (0.011) & 0.896 (0.009) & 0.896 (0.008) & \cellcolor{Ivory}{\bf0.908} (0.007)\\
    \bottomrule
    \end{tabular}
    \vspace{2mm}
    \caption{\small{Predictive performance (AUC-ROC) of different competing methods for the four primary outcome prediction tasks under consideration with respect to different levels of data-scarcity.}
    \label{tab:four-auc}}
    \end{table}

\begin{table}[t]
    \centering
    \footnotesize
    \renewcommand\arraystretch{1.2}
    \begin{tabular}{crcccc}
    \toprule 
    \rowcolor{GhostWhite}
    {\bf{Task}} & {\bf{Data}} & {\bf{Supervised}} & {\bf{Pretrain  (All)}} & {\bf{CoTrain}} & {\bf{AutoSelect}}\\
    \midrule 
   \parbox[h]{2mm}{\multirow{3}{*}{\rotatebox[origin=c]{90}{Mortality}}} 
    & 1\% & 0.047 (0.004) & 0.073 (0.007) & 0.056 (0.007) & \cellcolor{Ivory}{\bf 0.097} (0.009)\\
    & 10\% &0.127 (0.011) & 0.154 (0.015) &0.135 (0.009) & \cellcolor{Ivory}{\bf 0.168} (0.024)\\
    & 100\% & 0.267  (0.028) & 0.268 (0.019) & 0.262 (0.028) & \cellcolor{Ivory}{\bf 0.291} (0.033)\\
    \midrule 
   \parbox[h]{2mm}{\multirow{3}{*}{\rotatebox[origin=c]{90}{Shock}}} 
    & 1\% & 0.035 (0.005) & 0.052 (0.015 & 0.054 (0.019) & \cellcolor{Ivory}{\bf0.080} (0.019)\\
    & 10\% & 0.086 (0.025) & 0.096 (0.029) & 0.082 (0.021) & \cellcolor{Ivory}{\bf0.109} (0.031)\\
    & 100\% & 0.109 (0.031) & 0.206 (0.052) & 0.203 (0.057) & \cellcolor{Ivory}{\bf0.206 } (0.050)\\
    \midrule
   \parbox[h]{2mm}{\multirow{3}{*}{\rotatebox[origin=c]{90}{Kidney F.}}}
    & 1\% & 0.047 (0.002) & 0.074 (0.013) & 0.075 (0.011) & \cellcolor{Ivory}{\bf0.105} (0.009)\\
    & 10\% & 0.126 (0.014) & 0.113 (0.013) & 0.115 (0.014) & \cellcolor{Ivory}{\bf0.144} (0.014)\\
    & 100\% & 0.204 (0.017) & 0.223 (0.012) & 0.226 (0.016) & \cellcolor{Ivory}{\bf0.249} (0.017)\\
    \midrule
   \parbox[h]{2mm}{\multirow{3}{*}{\rotatebox[origin=c]{90}{Liver F.}}}
    & 1\% & 0.040 (0.004) & 0.076 (0.016) & 0.046 (0.008) & \cellcolor{Ivory}{\bf0.085} (0.011)\\
    & 10\% & 0.120 (0.015) & 0.107 (0.011) & 0.122 (0.015) & \cellcolor{Ivory}{\bf0.156} (0.021)\\
    & 100\% & 0.229 (0.018) & 0.184 (0.012) &0.226 (0.015) & \cellcolor{Ivory}{\bf0.257} (0.015)\\
    \bottomrule
    \end{tabular}
    \vspace{2mm}
    \caption{\small{Predictive performance (AUC-PR) of different competing methods for the four primary outcome prediction tasks under consideration with respect to different levels of data-scarcity.}
    \label{tab:four-ap}}
    \end{table}

\subsection{Additional Experiment Results}

Table~\ref{tab:four-auc} provides predictive performance (AUC-ROC) over the additional Liver Failure prediction task along with the other three tasks. We also report the AUC-PR (i.e., Average Precision (AP)) of different competing methods for the four primary outcome prediction tasks under consideration in Table~\ref{tab:four-ap}. As we could see, \autoselect outperforms both `Pretrain (All)' and `CoTrain' by a significantly large margin in low-data scenarios for all four tasks with respect to both metrics. 

\subsection{Top Tasks}

The features that are associated with the top trajectory forecast auxiliary tasks for each primary task are listed in Table~\ref{tab:top_tasks}. 

\begin{table}[h!]
\centering
\small
\renewcommand\tabcolsep{2pt}
\caption{Top Trajectory Forecast Tasks.\label{tab:top_tasks}}
\begin{tabular}{cp{12cm}} 
\toprule 
\rowcolor{GhostWhite}
{\bf Task} & {\bf Top Features} \\
\midrule                
Mortality & Invasive/Non-invasive blood pressure (diastolic, mean, systolic), anion gap,  heart rate, respiratory rate, minute ventilation obs, urine output foley, temperature, o2 saturation p, o2 flow, magnesium, sodium, glucose poc, calcium, glucose, hemoglobin, paw, arterial base excess, mean chc, arterial bicarbonate, base excess, rbc.\\ 
\midrule
Shock &	  Invasive/Non-invasive blood pressure (diastolic, mean, systolic),  heart rate, respiratory rate, urine output foley, temperature, o2 saturation p, o2 flow, magnesium, sodium, glucose, hemoglobin, paw.\\
\midrule
Kidney Failure & Invasive/Non-invasive blood pressure (diastolic, mean, systolic),  heart rate, respiratory rate, urine output foley, temperature, o2 saturation p,  magnesium, sodium, glucose. \\
\midrule
Liver Failure &  Invasive/Non-invasive blood pressure (diastolic, mean, systolic), anion gap,  heart rate, respiratory rate, minute ventilation obs, urine output foley, temperature, o2 saturation p, magnesium, sodium, glucose poc, calcium, glucose, hemoglobin, mean chc.\\
\bottomrule
\end{tabular}
\end{table}

\begin{scriptsize}
\begin{longtabu}{p{3cm}p{3cm}p{1cm}p{1.2cm}p{2.3cm}p{1cm}}\caption{List of input features used in the model.} 
\label{tab:features} \\ \toprule
Harmonized Name&
Display Names&
MIMIC code&
LOINC code&
Units&
Count
\\\midrule
\endhead
\bottomrule
\endfoot
Access pressure @mmHg & Access Pressure, Access mmHg & 224149, 29 &  & mmHg & 11490 \\ 
Albumin @G\_PER\_DL & Albumin, Albumin  (3.9-4.8), Albumin (>3.2) & 1521, 227456, 3727, 50862, 772 & 1751-7 & G\_PER\_DL, null & 161658 \\ 
Alt @IU\_PER\_L & ALT, Alanine Aminotransferase (ALT) & 220644, 50861, 769 & 1742-6 & IU\_PER\_L, null & 244169 \\ 
Anion gap @MEQ\_PER\_L & Anion gap, Anion Gap   (8-20), Anion Gap & 227073, 3732, 50868 & 1863-0 & MEQ\_PER\_L, null & 869575 \\ 
Ap @IU\_PER\_L & Alkaline Phosphate, Alkaline Phosphatase, Alk. Phosphate & 225612, 3728, 50863, 773 & 6768-6 & null, IU\_PER\_L & 232122 \\ 
Arterial base excess @MEQ\_PER\_L & Arterial Base Excess & 224828, 776 &  & null, MEQ\_PER\_L & 271605 \\ 
Arterial bicarbonate @MEQ\_PER\_L & TCO2 (calc) Arterial, Arterial CO2(Calc) & 225698, 777 &  & MEQ\_PER\_L & 406734 \\ 
Arterial pco2 @mmHg & Arterial CO2 Pressure, Arterial PaCO2 & 220235, 778 &  & mmHg & 406498 \\ 
Arterial ph @PH & Art.pH, PH (Arterial), pH (Art), Arterial pH & 1126, 223830, 4753, 780 &  & PH, null & 430827 \\ 
Arterial po2 @mmHg & Arterial O2 pressure, Arterial PaO2 & 220224, 779 &  & mmHg & 406341 \\ 
Ast @IU\_PER\_L & AST, Asparate Aminotransferase (AST) & 220587, 50878, 770 & 1920-8 & null, IU\_PER\_L & 244187 \\ 
Base excess @MEQ\_PER\_L & Base Excess & 50802 & 11555-0 & null, MEQ\_PER\_L & 329827 \\ 
Basophils @PERCENT & Basophils & 51146 & 704-7 & null, PERCENT & 172039 \\ 
Blood flow @ML\_PER\_MIN & Blood Flow (ml/min), Blood Flow ml/min & 224144, 79 &  & ML\_PER\_MIN & 116654 \\ 
Bp diastolic invasive @mmHg & Arterial Blood Pressure diastolic, ART BP Diastolic, Arterial BP [Diastolic], Arterial BP \#2 [Diastolic] & 220051, 225310, 8368, 8555 &  & mmHg & 3292274 \\ 
Bp diastolic non invasive @mmHg & Non Invasive Blood Pressure diastolic, Manual Blood Pressure Diastolic Left, Manual Blood Pressure Diastolic Right, Manual BP [Diastolic], NBP [Diastolic] & 220180, 224643, 227242, 8440, 8441 &  & mmHg & 2863944 \\ 
Bp map invasive @mmHg & Arterial Blood Pressure mean, ART BP mean, Arterial BP Mean, Arterial BP Mean \#2, Arterial Mean \#3 & 220052, 225312, 52, 6702, 6927 &  & mmHg & 3280702 \\ 
Bp mean non invasive @mmHg & Non Invasive Blood Pressure mean, Manual BP Mean(calc), NBP Mean & 220181, 443, 456 &  & mmHg & 2843856 \\ 
Bp systolic invasive @mmHg & Arterial Blood Pressure systolic, ART BP Systolic, Arterial BP [Systolic], Arterial BP \#2 [Systolic] & 220050, 225309, 51, 6701 &  & mmHg & 3293052 \\ 
Bp systolic non invasive @mmHg & Non Invasive Blood Pressure systolic, Manual Blood Pressure Systolic Left, Manual Blood Pressure Systolic Right, Manual BP [Systolic], NBP [Systolic] & 220179, 224167, 227243, 442, 455 &  & mmHg & 2865402 \\ 
Bun @MG\_PER\_DL & BUN, BUN    (6-20), Urea Nitrogen, BUN (6-20) & 1162, 225624, 3737, 51006, 781 & 3094-0 & MG\_PER\_DL, null & 893969 \\ 
Calcium @ & Calcium, Calcium non-ionized, Calcium   (8.8-10.8), Calcium, Total, Calcium (8.4-10.2) & 1522, 225625, 3746, 50893, 786 & 2000-8 & MEQ\_PER\_L, null & 323 \\ 
Calcium @MEQ\_PER\_L & Calcium, Calcium non-ionized, Calcium   (8.8-10.8), Calcium, Total, Calcium (8.4-10.2) & 1522, 225625, 3746, 50893, 786 & 2000-8 & MEQ\_PER\_L, null & 681992 \\ 
Cardiac index @UNKNOWN\_UOM & Cardiac Index, Cardiac Index (CI NICOM) & 116, 228368 &  & UNKNOWN\_UOM, L\_PER\_MIN\_PER\_M2 & 209483 \\ 
Cardiac output rate @L\_PER\_MIN & Cardiac Output (thermodilution), Cardiac Output (CCO), CO (Arterial), CO (PiCCO), Cardiac Output (CO NICOM), C.O. (fick), C.O.(thermodilution) & 220088, 224842, 227543, 228178, 228369, 89, 90 &  & L\_PER\_MIN & 294655 \\ 
Chloride @MEQ\_PER\_L & Chloride, Chloride (serum), Chloride (whole blood), Chloride  (100-112), Chloride, Whole Blood, Chloride (100-112) & 1523, 220602, 226536, 3747, 50806, 50902, 788 & 2069-3, 2075-0 & MEQ\_PER\_L, null & 959368 \\ 
Co2 @MEQ\_PER\_L & HCO3 (serum), Calculated Bicarbonate, Whole Blood, Bicarbonate, Carbon Dioxide, HCO3 & 227443, 50803, 50882, 787, 812 & 1959-6, 1963-8 & null, MEQ\_PER\_L & 891999 \\ 
Creatine kinase @IU\_PER\_L & CK (CPK), Creatine Kinase (CK), CPK & 225634, 50910, 784 & 2157-6 & IU\_PER\_L, null & 149871 \\ 
Creatinine @MG\_PER\_DL & Creatinine, Creatinine   (0-0.7), Creatinine (0-1.3) & 1525, 220615, 3750, 50912, 791 & 2160-0 & MG\_PER\_DL, null & 899717 \\ 
Creatinine kinase mb @NG\_PER\_ML & CK-MB, Creatine Kinase, MB Isoenzyme, CPK/MB & 227445, 50911, 785 & 6773-6 & NG\_PER\_ML, null & 89737 \\ 
Cvp @mmHg & CVP, Central Venous Pressure & 113, 220074 &  & mmHg, PERCENT & 1680374 \\ 
Eosinophils @PERCENT & Eosinophils & 51200 & 711-2 & null, PERCENT & 172042 \\ 
Exhaled minute ventilation low @ & High exhaled min vol, high exhaled min vol, HIGH EXHALED MIN VOL, High Exhaled min vol, high exhaled min.vol, High exhaled min.vol, high exhale MV, HIGH EXHALE MV, High exhaled MV, High exhale MV, high exhale mv, Low Exhaled Min Vol, High Exhaled Min vol & 1010, 1102, 1223, 1313, 1323, 1380, 1720, 1723, 1724, 2123, 2127, 434, 5744 &  & L\_PER\_MIN, null & 186 \\ 
Exhaled minute ventilation low @L\_PER\_MIN & High exhaled min vol, high exhaled min vol, HIGH EXHALED MIN VOL, High Exhaled min vol, high exhaled min.vol, High exhaled min.vol, high exhale MV, HIGH EXHALE MV, High exhaled MV, High exhale MV, high exhale mv, Low Exhaled Min Vol, High Exhaled Min vol & 1010, 1102, 1223, 1313, 1323, 1380, 1720, 1723, 1724, 2123, 2127, 434, 5744 &  & L\_PER\_MIN, null & 386163 \\ 
Expiratory ratio @RATIO & Expiratory Ratio & 226871 &  & RATIO & 207217 \\ 
Fio2 analyzed @TORR & FiO2 (Analyzed) & 189 &  & TORR & 6914 \\ 
Glucose @MG\_PER\_DL & Glucose, Glucose (serum), Glucose (whole blood), Glucose (70-105) & 1529, 220621, 226537, 50809, 50931, 811 & 2339-0, 2345-7 & MG\_PER\_DL, null & 1172288 \\ 
Glucose poc @MG\_PER\_DL & Glucose finger stick, Fingerstick Glucose & 225664, 807 &  & MG\_PER\_DL & 678665 \\ 
Heart rate @BPM & Heart Rate & 211, 220045 &  & BPM & 7938853 \\ 
Hematocrit @PERCENT & Hematocrit (serum), Hematocrit (whole blood - calc), Hematocrit (35-51), Hematocrit, Calculated, Hematocrit & 220545, 226540, 3761, 50810, 51221, 813 & 20570-8, 4544-3 & null, PERCENT & 1094791 \\ 
Hemoglobin @G\_PER\_DL & Hemoglobin, HGB  (10.8-15.8), Absolute Hemoglobin & 220228, 3759, 50811, 50855, 51222, 814 & 718-7 & G\_PER\_DL, null & 942483 \\ 
Inr @RATIO & INR, INR(PT), INR (2-4 ref. range) & 1530, 227467, 51237, 815 & 5895-7 & RATIO, null & 161338 \\ 
Inr @ & INR, INR(PT), INR (2-4 ref. range) & 1530, 227467, 51237, 815 & 5895-7 & RATIO, null & 371373 \\ 
Insp pressure @CM\_H2O & High Insp. Pressure, low insp pressure, Low Insp. Pressure, Low insp pressure, low IP & 218, 3143, 436, 6864, 7094 &  & CM\_H2O & 384580 \\ 
Insp time @S & Insp. Time, Inspiratory Time, Insp.Time, Insp time & 1655, 2000, 224738, 3009, 6315 &  & null, S & 364845 \\ 
Inspiratory ratio @RATIO & Inspiratory Ratio & 226873 &  & RATIO & 206635 \\ 
Ionized calcium @MEQ\_PER\_L & Ionized Calcium, Free Calcium & 225667, 50808, 816 & 1994-3 & null, MEQ\_PER\_L & 303125 \\ 
Ketones urine @MG\_PER\_DL & Ketone & 51484 & 5797-6 & MG\_PER\_DL & 10617 \\ 
Lactate @MMOL\_PER\_L & Lactic Acid, Lactate, Lactic Acid(0.5-2.0) & 1531, 225668, 50813, 818 & 32693-4 & MMOL\_PER\_L, null & 233021 \\ 
Lymphocytes diff @PERCENT & Differential-Lymphs & 225641, 798 &  & PERCENT & 41425 \\ 
Magnesium @MG\_PER\_DL & Magnesium, Magnesium (1.6-2.6) & 1532, 220635, 50960, 821 & 2601-3 & MG\_PER\_DL, null & 762825 \\ 
Mean ch @PG & MCH & 51248 & 785-6 & null, PG & 747387 \\ 
Mean chc @PERCENT & MCHC & 51249 & 786-4 & null, PERCENT & 747756 \\ 
Mean cv @FL & MCV & 51250 & 787-2 & null, FL & 747377 \\ 
Minute ventilation obs @L\_PER\_MIN & Minute Volume, Minute Volume(Obser) & 224687, 448, 450 &  & L\_PER\_MIN & 851154 \\ 
Monocytes @PERCENT & Monocytes & 51254 & 742-7 & null, PERCENT & 172044 \\ 
Neutrophils urine @PERCENT & Neutrophils & 51256 & 761-7 & null, PERCENT & 170591 \\ 
O2 flow @L\_PER\_MIN & O2 Flow, O2 Flow (additional cannula), O2 Flow (lpm), O2 Flow (lpm) \#2 & 223834, 227287, 470, 471, 50815 & 3151-8 & L\_PER\_MIN & 622606 \\ 
O2 saturation @PERCENT & Arterial O2 Saturation, Oxygen Saturation, SaO2 & 220227, 50817, 834 & 20564-1 & PERCENT, null & 1828799 \\ 
O2 saturation p @PERCENT & O2 saturation pulseoxymetry, SpO2, SpO2-L & 220277, 646, 6719 &  & PERCENT & 6086176 \\ 
P co2 @mmHg & pCO2 & 50818 & 11557-6 & null, mmHg & 490470 \\ 
P o2 @mmHg & pO2 & 50821 & 11556-8 & null, mmHg & 490481 \\ 
Paw @CM\_H2O & MEAN AIRWAY PRESS, PAW, Paw High, Mean Airway Pressure, Mean PAW, Mean PAW [Meas] & 1672, 2229, 223873, 224697, 3502, 3503, 444 &  & null, CM\_H2O & 1101930 \\ 
Peep observed @CM\_H2O & PEEP, MEASURED PEEP & 505, 6924 &  & CM\_H2O, null & 350192 \\ 
Ph @PH & PH, pH & 1673, 50820 & 11558-4 & null, PH & 530708 \\ 
Ph urine @PH & urine pH, urine ph, Urine pH, PH (dipstick), pH, urine PH & 1352, 1495, 1880, 220734, 51491, 6754 & 5803-2 & PH, null & 129352 \\ 
Phosphorous @MEQ\_PER\_L & Phosphorous, Phosphate, Phosphorous(2.7-4.5) & 1534, 225677, 50970, 827 & 2777-1 & null, MEQ\_PER\_L & 681128 \\ 
Plateau pressure @CM\_H2O & Plateau Pressure & 224696, 543 &  & CM\_H2O & 242730 \\ 
Platelet @ & Platelet Count, Platelet  (150-440), Platelets & 227457, 3789, 51265, 828 & 777-3 & K\_PER\_UL, null & 5313 \\ 
Platelet @K\_PER\_UL & Platelet Count, Platelet  (150-440), Platelets & 227457, 3789, 51265, 828 & 777-3 & K\_PER\_UL, null & 865223 \\ 
Potassium @MEQ\_PER\_L & Potassium, Potassium (serum), Potassium (whole blood), Potassium  (3.5-5.3), Potassium, Whole Blood, Potassium (3.5-5.3) & 1535, 227442, 227464, 3792, 50822, 50971, 829 & 2823-3, 6298-4 & MEQ\_PER\_L, null & 1181418 \\ 
Potassium urine @MEQ\_PER\_L & Potassium, Urine & 51097 & 2828-2 & null, MEQ\_PER\_L & 10850 \\ 
Protein urine @MG\_PER\_DL & Protein & 51492 & 5804-0 & null, MG\_PER\_DL & 36043 \\ 
Psv @UNKNOWN\_UOM & PSV Level, Pressure Support, pressure support, PSV & 224701, 578, 7332, 7595 &  & CM\_H2O, null, UNKNOWN\_UOM & 406535 \\ 
Pt @S & PT, Prothrombin time, ProTime, PT(11-13.5) & 1286, 227465, 3793, 51274, 824 & 5902-2 & S, null & 530740 \\ 
Ptt @S & PTT, Ptt, PTT(22-35) & 1533, 227466, 3796, 51275, 825 & 3173-2 & S, null & 539725 \\ 
Rbc @PER\_UL & Red Blood C(3.6-6.2), RBC(3.6-6.2), Red Blood Cells, RBC & 3799, 4197, 51279, 833 & 789-8 & PER\_UL, null & 748036 \\ 
Rdw @PERCENT & RDW & 51277 & 788-0 & null, PERCENT & 746239 \\ 
Replacement rate @ML\_PER\_H & Replacement Rate, Replace Rate ml/hr & 224153, 611 &  & ML\_PER\_H & 118496 \\ 
Respiratory rate @BREATHS\_PER\_MIN & High Resp. Rate, Respiratory Rate, high rr, Resp Rate & 219, 220210, 3142, 3603, 618 &  & BREATHS\_PER\_MIN & 7904015 \\ 
Respiratory rate spont @BREATHS\_PER\_MIN & Spont Resp rate, Spont RR, Respiratory Rate (spontaneous), Resp Rate (Spont), Spon RR (Mech.), Spont. Resp. Rate & 1884, 224422, 224689, 614, 651, 653 &  & null, BPM, BREATHS\_PER\_MIN & 714134 \\ 
Respiratory rate total @BPM & Respiratory Rate (Total), Resp Rate (Total) & 224690, 615 &  & BREATHS\_PER\_MIN, BPM & 812342 \\ 
Rrt output @ML & Ultrafiltrate Output, Ultrafiltrate Ultrafiltrate, dialysis output, dialysis out, Dialysis out, DIALYSIS OUT, dialysis, DIALYSIS, Dialysis, PD dialysate out, Dialysis Out, dialysate out, Dialysis output, ULTRAFILTRATE, ultrafiltrate in, peritoneal dialysis, ultrafiltrate, ultrafiltrate out & 226457, 40286, 40425, 40426, 40507, 40613, 40624, 40690, 40745, 40789, 41374, 41623, 42536, 43703, 44349, 44843, 44890, 46622 &  & ML, null & 123788 \\ 
Sodium @MEQ\_PER\_L & Sodium, Sodium (serum), Sodium (whole blood), Sodium  (135-148), Sodium, Whole Blood, Sodium (135-148) & 1536, 220645, 226534, 3803, 50824, 50983, 837 & 2947-0, 2951-2 & MEQ\_PER\_L, null & 1000050 \\ 
Svo2 @PERCENT & SVO2, SvO2, Mixed Venous O2\% Sat, svo2, Swan SVO2, SV02, svo2 RIJ & 2194, 223772, 225674, 2669, 664, 7361, 8186, 823, 838 &  & PERCENT & 250574 \\ 
Tbili @MG\_PER\_DL & Total Bili, Total Bilirubin, Bilirubin, Total, Total Bili (0-1.5) & 1538, 225690, 50885, 848 & 1975-2 & MG\_PER\_DL, null & 263182 \\ 
Tco2 @MEQ\_PER\_L & TCO2        (21-30), Calculated Total CO2 & 3808, 50804 & 34728-6 & null, MEQ\_PER\_L & 490499 \\ 
Temperature @CEL & Temperature Fahrenheit, Temperature Celsius, Temp Axillary [F], Temp Rectal [F], Temperature C, Temperature F & 223761, 223762, 3652, 3654, 676, 678 &  & CEL & 2203935 \\ 
Troponin t @NG\_PER\_ML & Troponin-T, Troponin T & 227429, 51003 & 6598-7 & NG\_PER\_ML & 79888 \\ 
Urine output @ML & Void, OR Urine, PACU Urine, Drainage Bag, OR Out OR Urine, OR Out PACU Urine, Urine Out Void, Urine Out Incontinent, PACU Out PACU Urine, Urine Out Other, ER URINE, TRUE URINE, Urine ., Dialysis indwelling, True Urine & 226560, 226627, 226631, 227701, 40061, 40065, 40069, 40085, 40288, 40405, 42001, 42507, 43175, 44286, 45927 &  & ML, null & 252169 \\ 
Urine output foley @ML & Foley, Urine Out Foley & 226559, 40055 &  & ML, null & 3093578 \\ 
Vt obs @ML\_PER\_BREATH & Tidal Volume (observed), tidal volumes, tidal vol, tidal volume, Tidal Volume, Tidal Volume (Obser) & 224685, 2400, 2408, 2534, 681, 682 &  & ML\_PER\_BREATH & 701861 \\ 
Vt spont @ML\_PER\_BREATH & Spont Vt, Tidal Volume (spontaneous), svt, Spontaneous VT, spontaneous VT, spont tidal volumes, spont Tidal volumes, spont Vt's, Spon. Vt (L) (Mech.), Spont. Tidal Volume, Tidal Volume (Spont) & 224421, 224686, 2553, 2566, 3004, 3050, 3083, 3086, 652, 654, 684 &  & ML\_PER\_BREATH, null & 494654 \\ 
Wbc count @K\_PER\_UL & WBC   (4-11,000), WBC, WBC 4.0-11.0, White Blood Cells, WBC (4-11,000) & 1127, 1542, 220546, 4200, 51301, 861 & 804-5 & K\_PER\_UL, null & 842129 \\ 
Weight @KG & Daily Weight & 224639, 763 &  & KG & 93386 \\ 
\end{longtabu}
\end{scriptsize}

\end{document}